\documentclass[]{fairmeta}
\usepackage{amsmath,amsfonts,bm}

\def\eqref#1{equation~\ref{#1}}
\def\Eqref#1{Equation~\ref{#1}}
\def\1{\bm{1}}

\DeclareMathAlphabet{\mathsfit}{\encodingdefault}{\sfdefault}{m}{sl}
\SetMathAlphabet{\mathsfit}{bold}{\encodingdefault}{\sfdefault}{bx}{n}

\usepackage{hyperref}
\usepackage{url}

\usepackage{booktabs}
\usepackage{makecell}
\usepackage{graphicx}
\usepackage{float}
\usepackage{multirow}

\usepackage{algorithm}         % floating 'Algorithm' environment
\usepackage{algpseudocode}     % algorithmicx pseudocode (\State, \If, ...)
\usepackage{xspace}            % lets \kb, \kite behave like words

\usepackage[most]{tcolorbox}
\tcbset{enhanced, sharp corners, boxrule=0.3pt, left=6pt,right=6pt,top=4pt,bottom=4pt}

\newtcolorbox{SystemMsg}{colback=gray!15,title={System}}
\newtcolorbox{UserMsg}{colback=blue!5,title={User}}
\newtcolorbox{AssistantMsg}{colback=green!5,title={Assistant}}

\usepackage{listings}
\tcbuselibrary{listings,breakable}
\usepackage{xparse}

\lstdefinestyle{promptstyle}{
  basicstyle=\ttfamily\small,
  breaklines=true,
  breakatwhitespace=false,   % allow breaks anywhere if needed
  breakautoindent=false,     % <-- don't inherit the current indent
  breakindent=0pt,           % <-- no extra hanging indent
  columns=fullflexible,      % saner spacing for long lines
  keepspaces=true,           % keep literal spaces (no magic stretching)
  tabsize=2,                 % tabs won't jump too far
  numbers=left,
  numbersep=6pt,
  showstringspaces=false,
}

\newcommand{\rolelabel}[1]{\textsf{\textbf{#1}}}

\newtcblisting{PromptBox}[2][]{%
  enhanced,
  breakable,
  sharp corners,
  boxrule=0.4pt,
  colback=gray!2,
  colframe=black!12,
  left=1em,right=1em,top=0.75em,bottom=0.75em,
  listing only,
  listing options={style=promptstyle},
  title={\rolelabel{#2}\IfValueT{#1}{\hfill\footnotesize\emph{#1}}},
}

\usepackage{listings} % arXiv-safe (use minted if allowed)
\usepackage{wrapfig}
\usepackage[font=small,labelfont=bf]{caption} % tighter caption

\newcommand{\stark}{\texttt{STARK}\xspace}

\title{STARK: Strategic Team of Agents for Refining Kernels}

\author[1,2,*]{Juncheng Dong}
\author[1]{Yang Yang}
\author[1]{Tao Liu}
\author[1]{Yang Wang}
\author[1]{Feng Qi}
\author[2]{Vahid Tarokh}
\author[1]{Kaushik Rangadurai}
\author[1]{Shuang Yang}

\affiliation[1]{Meta Ranking AI Research}
\affiliation[2]{Duke University}

\contribution[*]{Work done during an internship at Meta}
\abstract{
The efficiency of GPU kernels is central to the progress of modern AI, yet optimizing them remains a difficult and labor-intensive task due to complex interactions between memory hierarchies, thread scheduling, and hardware-specific characteristics. While recent advances in large language models (LLMs) provide new opportunities for automated code generation, existing approaches largely treat LLMs as single-shot generators or naive refinement tools, limiting their effectiveness in navigating the irregular kernel optimization landscape. We introduce an LLM agentic framework for GPU kernel optimization that systematically explores the design space through multi-agent collaboration, grounded instruction, dynamic context management, and strategic search. This framework mimics the workflow of expert engineers, enabling LLMs to reason about hardware trade-offs, incorporate profiling feedback, and refine kernels iteratively. We evaluate our approach on KernelBench, a benchmark for LLM-based kernel optimization, and demonstrate substantial improvements over baseline agents: our system produces correct solutions where baselines often fail, and achieves kernels with up to 16$\times$ faster runtime performance. These results highlight the potential of agentic LLM frameworks to advance fully automated, scalable GPU kernel optimization.
}

\date{\today}
\correspondence{\email{yzyang@meta.com}, \email{shuangyang@meta.com}}

\begin{document}

\maketitle

\section{Introduction}

\begin{wrapfigure}{r}{0.50\linewidth}
  \vspace{-8pt}
  \centering
  \includegraphics[width=\linewidth]{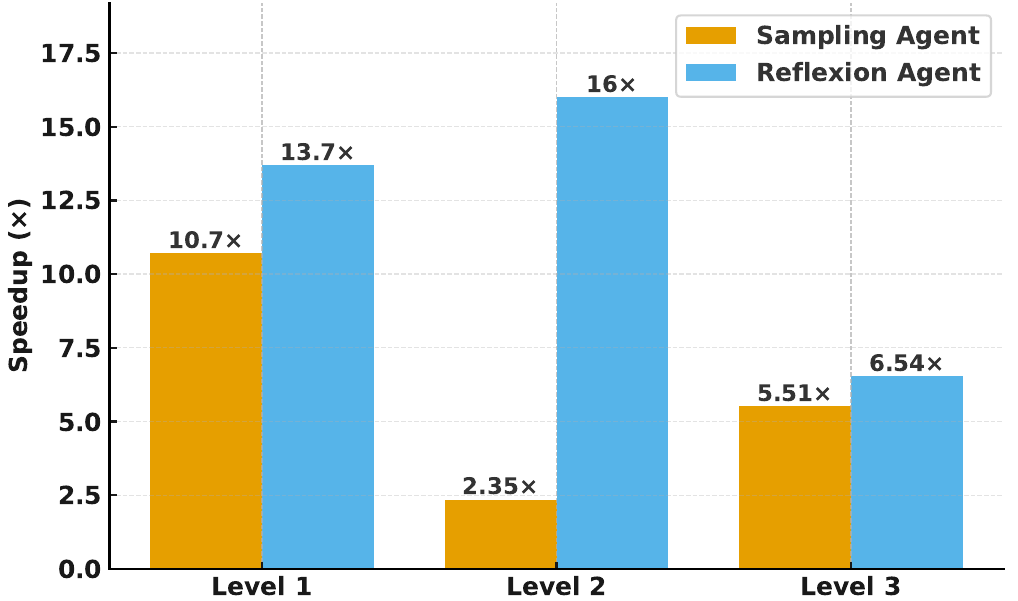}
  \vspace{-6pt}
  \caption{\textbf{Speedup of \stark over baseline agents on KernelBench (L1–L3) with same number of attempts}.
  Bars report GPU wall-clock speedups ($\times$) relative to the \emph{Sampling} and \emph{Reflexion} agents; higher is better. \stark reaches up to $16\times$ over Reflexion (L2) and $10.7\times$ over Sampling (L1).}
  \label{fig:agent_speedups}
  \vspace{-15pt}
\end{wrapfigure}

Artificial intelligence (AI) has advanced at an unprecedented pace, transforming both research and real-world applications. While innovations in model architectures and training algorithms have been central to this progress, the efficiency of the computational infrastructure that executes them is equally critical. 
% As the engines driving today’s AI revolution, graphics processing units (GPUs) dictate the scalability and performance of modern systems.
At the core of modern AI systems are \emph{GPU kernels}, which implement fundamental operations such as matrix multiplication and convolution. 
Even modest improvements in GPU kernel efficiency can translate into significant reductions in training time, inference latency, and deployment cost, making kernel optimization a cornerstone for sustaining AI’s rapid growth.
% Notably, the full potential of state-of-the-art AI and ML architectures can only be realized when these kernels are carefully optimized for speed and scalability. 

Despite their importance, designing and optimizing GPU kernels remains a major challenge. The performance of a kernel depends on subtle interactions between thread scheduling, memory hierarchy utilization, synchronization, and hardware-specific characteristics. Small changes in tiling strategies, loop unrolling, or memory alignment can yield disproportionate effects on runtime. As a result, the kernel optimization landscape is highly irregular, architecture-dependent, and difficult to navigate. Existing approaches largely fall into two categories: \emph{manual optimization} by expert engineers, which is effective but labor-intensive and difficult to scale; and \emph{automated compilers and domain-specific languages} (DSLs) such as TVM and Triton~\citep{chen2018tvm,tillet2019triton}, which apply heuristics or search but often struggle with irregular operators and hardware variability~\citep{zheng2020ansor,zheng2020flextensor}. 

The rapid progress of large language models (LLMs) opens a new opportunity for kernel optimization. Beyond their ability to generate correct code, LLMs can be guided to reason about hardware trade-offs, adapt to profiling feedback, and iteratively refine implementations. However, prior work has mostly treated LLMs as single-shot code generators or simple refinement tools~\citep{ouyang2025kernelbench}, which underutilizes their potential for structured exploration of the kernel design space. 
To build a more powerful agent, we identify and address three critical limitations in existing methods:
\begin{enumerate}
    \item \textbf{Naive exploration strategy.} Current agents typically refine code linearly, learning only from the immediately preceding attempt. This simplistic process neglects the rich history of prior attempts and fails to effectively balance the exploration-exploitation trade-off, often getting trapped in local optima.

    \item \textbf{Monolithic agent design.} Kernel optimization is a multifaceted task requiring distinct capabilities for planning, implementation, and reflection. By assigning all these responsibilities to a single, generalist LLM, current agents operate inefficiently.

    \item \textbf{Planning-implementation gap.} We observe a failure mode particularly acute in this domain: LLMs frequently devise a correct high-level optimization plan (e.g., ``apply memory tiling'') but fail to translate it into valid low-level CUDA code. This gap stems from the relative scarcity of expert-level kernel code in the models' training data.
\end{enumerate}

To address these limitations, we introduce \stark (\emph{Strategic Team of Agents for Refining Kernels}), a novel framework for automated GPU-kernel optimization. Our contributions are threefold:
\begin{itemize}
    \item \textbf{Collaborative multi-agent workflow.} We design a workflow with specialized agents for planning, coding, and reflection, mirroring an expert development cycle and overcoming the inefficiencies of monolithic designs.
    \item \textbf{Bridging the planning--implementation gap.} We propose two mechanisms---\emph{grounded instruction} and \emph{dynamic context windows}---that translate high-level strategies into precise, actionable code edits, ensuring robust coordination across agents.
    \item \textbf{Strategic search for refinement.} We incorporate a search policy that balances exploration and exploitation over prior attempts, enabling systematic discovery of strong kernels.
\end{itemize}

\begin{figure}[h]
    \centering
    \includegraphics[width=1.0\linewidth]{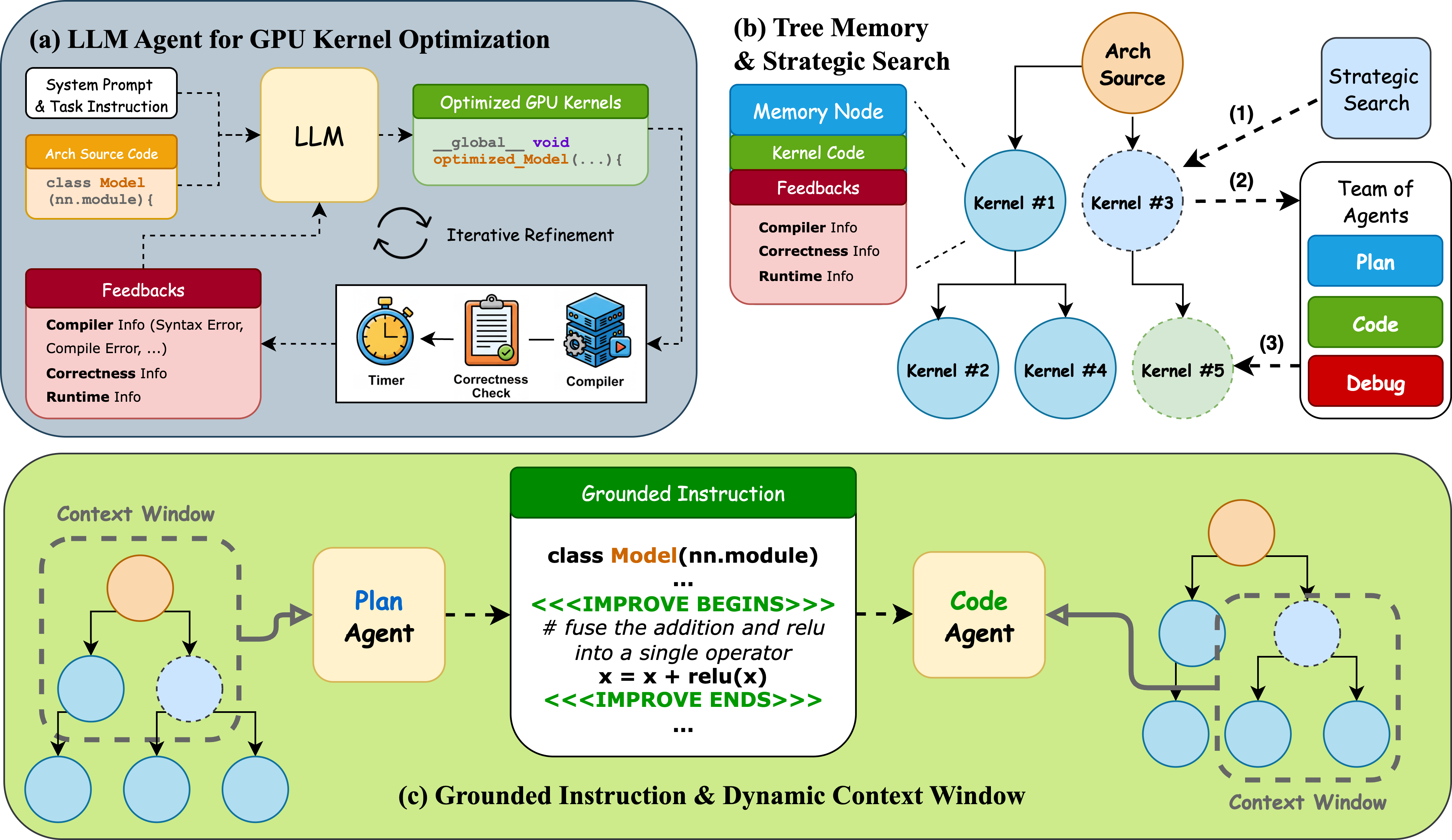}
    \caption{\textbf{Overview of \stark}. \textbf{(a)} Prior LLM-based kernel optimizers rely on a monolithic agent with local iterative refinement. \textbf{(b)} \stark replaces this with a collaborative multi-agent workflow (plan/code/debug) coupled with strategic search over a tree memory. \textbf{(c)} The plan agent issues \emph{grounded instructions} that anchor edits to code spans; \emph{dynamic context windows} surface role-specific history; and the debug agent repairs failures. See Section~\ref{sec:stark} for details.}
    \label{fig:intro_fig}
\end{figure}

% Together, these components enable systematic and scalable exploration of the vast GPU kernel search space. 
We evaluate our framework on \textbf{KernelBench}~\citep{ouyang2025kernelbench}, a benchmark designed to assess LLM-based GPU kernel optimization. Experiments show that combining these improvements leads to an agent system significantly more competitive than the baseline agents in both runtime performance and success rate across diverse kernel problems, authoring competitive kernels for the challenging problems in KernelBench where the baseline agents struggle to even find a working solution.
Notably, \stark achieves more than $10\times$ speedup over kernels produced by the baseline agents (i.e., the optimized kernels run in under one-tenth the time of the baseline.).
Overall, our work suggests that LLM-driven agents represent a promising step toward fully automated GPU kernel optimization.

\section{Related Work}
\label{sec:related_work}
We only review the most relevant prior work here and defer the complete discussion to Appendix~\ref{app:complete_related_work}.

The optimization of GPU kernels has progressed from empirical auto-tuning frameworks that perform black-box parameter searches~\citep{vanwerkhoven2019kerneltuner, nugteren2015cltune} and compiler-based approaches with static heuristics~\citep{yang2010gpgpu}, to the use of machine learning (ML). ML-based techniques have been used to replace hand-tuned heuristics in production compilers~\citep{trofin2021mlgo}, learn cost models to guide optimization~\citep{chen2018tvm}, and even learn directly from raw source code without manual feature engineering~\citep{cummins2017endtoend}. 
A significant leap was the use of deep reinforcement learning to discover fundamentally new algorithms, as demonstrated by AlphaTensor's success in finding faster matrix multiplication methods~\citep{fawzi2022alphatensor}. While powerful, these prior works either optimize within a fixed search space or operate in purely formal domains. Our work addresses these limitations by operating directly on source code to implement novel, structural changes.

The emergence of powerful Large Language Models (LLMs) has revolutionized programmatic interaction with source code, demonstrating a remarkable proficiency in generating code for diverse applications from competitive programming to compiler testing~\citep{gu2023llm,zhong2024can,jain2025livecodebench}. This capability has catalyzed a paradigm shift away from single-shot code generation and toward the development of autonomous LLM agents. An agent enhances a base LLM with planning, memory, and tool-use capabilities to direct its own workflow~\citep{weng2023llmagents}. The success of frameworks like SWE-agent in independently resolving complex GitHub issues has validated the power of this approach for software engineering (SWE)~\citep{yang2024sweagent}.
While the application of LLM agents to SWE is a burgeoning field of research~\citep{yang2024sweagent,antoniades2024swe,yang2025swe}, their potential in the specialized domain of GPU kernel optimization remains largely unexplored. To fill this gap, we designed \stark, an agent framework with capabilities tailored to the unique challenges of this domain.

\section{Preliminary}

\subsection{LLMs and Autoregressive Generation}
Given an input sequence $x=(x_1,x_2,\dots,x_n)$ (e.g., the task instruction) as the context, an LLM $p_\theta$ with parameters $\theta$ generates an output sequence $y= (y_1,y_2,\dots,y_m)$ where $y_t \in \mathcal{Y},t\in\{1,\dots,m\}$ are tokens. 
Pretrained on a massive corpus of text, LLMs autoregressively generate the next token $y_t$ conditioning on $x$ and all the previously generated token $y_{<t} = (y_1,\dots,y_{t-1})$. 
Specifically, at each time $t$, the LLM first computes the logits $z_\theta(y|y_{<t},x)$ for each token $y$ in the vocabulary $\mathcal{Y}$ and generate $y_t$ following the conditional distribution 
\begin{equation}\label{eqn:next-token}
p_\theta(y_t|y_{<t},x)=\frac{\exp(z_\theta(y_t|y_{<t},x)/\tau)}{\sum_{y' \in \mathcal{Y}}\exp(z_\theta(y'|y_{<t},x)/\tau)}. 
\end{equation}
The temperature parameter $\tau > 0$ modulates the randomness of an LLM's output. 
Higher values of $\tau$ flatten the next token distribution in~\Eqref{eqn:next-token}, encouraging creative and diverse responses. Conversely, lower values sharpen the distribution, promoting deterministic and high-fidelity outputs.

This trade-off is critical in complex tasks, as different sub-problems demand different behaviors. For instance, \textbf{planning} and \textbf{exploration} benefit from a high temperature to generate novel strategies, whereas tasks requiring precision and factual correctness, such as \textbf{code implementation}, necessitate a low temperature to ensure reliability. A single agent with a fixed temperature is ill-equipped to handle this dichotomy. This observation is a core motivation for \stark's multi-agent design, which allows specialized agents to operate at distinct temperatures tailored to their roles, i.e., a high $\tau$ for the creative plan agent and a low $\tau$ for the precise code agent.

\subsection{KernelBench}

\textbf{KernelBench}~\citep{ouyang2025kernelbench} is a recently proposed benchmark specifically designed for assessing LLM-based GPU kernel optimization. Unlike prior evaluations that focus only on code correctness or small-scale operator tests, KernelBench provides a principled and reproducible testbed that measures both correctness and runtime efficiency across a broad spectrum of GPU workloads.
KernelBench comprises a suite of optimization tasks, categorized into three difficulty levels. For each task, the objective is to create a custom GPU kernel that is functionally equivalent to a provided PyTorch reference implementation while minimizing its wall-clock execution time. See an example of the KernelBench task in Appendix~\ref{app:kb-example}.

Specifically, \textbf{Level 1} tasks focus on single, common operators such as matrix multiplication and convolution, serving as a baseline for fundamental optimization capabilities; \textbf{Level 2} tasks comprise tasks with multiple operators fused into a single kernel, testing the ability to manage more complex dataflows and scheduling; \textbf{Level 3} tasks represent the highest difficulty, featuring popular full ML architectures such as the ResNet~\citep{he2016deep} and LSTM~\citep{hochreiter1997long}, which involve highly irregular computations and intricate memory access patterns that are challenging for both human experts and automated systems to optimize effectively.

\section{STARK: Strategic Team of Agents for Refining Kernels}\label{sec:stark}

\textbf{Framework Overview.} 
We now present \stark, an agentic framework for GPU-kernel optimization. \stark organizes kernel refinement into three layers: (i) a \emph{multi-agent workflow} that separates planning, coding, and debugging, (ii) \emph{coordination mechanisms} with \emph{grounded instruction} to anchor planned edits to concrete code spans and \emph{dynamic context windows} that surface role-specific history (e.g., prior attempts, failures, profiler feedback) to each agent, and (iii) a \emph{strategic search} policy that balances exploration and exploitation across iterative attempts. Notably, multi-agent workflow and grounded instruction improve reliability even under a single-attempt budget, whereas dynamic context windows and strategic search deliver most of their gains when multiple attempts are allowed. Figure~\ref{fig:intro_fig} provides an overview; the following subsections detail each component in turn.

\subsection{Multi-Agent Collaboration}
Optimizing GPU kernels is inherently multifaceted and mirrors expert team workflows. A single agent typically fails to balance correctness, performance, and exploration across a vast, irregular design space. In particular, \emph{strategy discovery} (e.g., fusion, vectorization, shared-memory tiling) benefits from higher-temperature generation that encourages diversity whereas \emph{strategy realization}, i.e., committing those ideas to code, requires low-temperature precision to avoid errors. We therefore adopt a multi-agent framework that enables role specialization through LLMs.

\textbf{Multi-Agent Design (MAD).} Specifically, \stark decomposes kernel optimization into three roles -- \emph{plan}, \emph{code}, and \emph{debug}. 
Using a role-specific context window (Section~\ref{sec:context_window}) with selected prior attempts and execution outcomes, the \textbf{plan} agent proposes targeted transformations to either the source kernel or a candidate chosen by the strategic search policy (Section~\ref{sec:search}), emitting \emph{grounded instructions} (Section~\ref{sec:grounded_instruction}) that anchor edits to explicit code spans. The \textbf{code} agent consumes grounded instructions and translates them into executable GPU-kernel code, conditioning on its own context window to improve adherence and code quality. The \textbf{debug} agent repairs promising but failing candidates by consulting the plan agent's instructions and compiler/runtime diagnostics, producing a working kernel that realizes the intended transformation.

\textbf{Benefits of MAD.} Role specialization lets each agent use prompts and base LLMs matched to its objective. In our instantiation, we choose \texttt{Claude Sonnet 4} with temperature $\tau{=}0.8$ for the plan agent to encourage strategy diversity, and the same model with $\tau{=}0.1$ for the code and debug agents to enforce precision. Despite this simple setup, MAD already performs strongly (see Section~\ref{sec:experiments}). We underscore that because the design is modular, we can swap in planners with richer kernel-optimization priors or code-specialist reasoning models to further improve results.
In addition, modularity also exposes bottlenecks. We observe that the dominant bottleneck is code-synthesis fidelity: LLMs often need multiple attempts to faithfully implement a given instruction. Finally, MAD makes targeted post-training straightforward: we can fine-tune the base LLM for a specific agent (e.g., the code agent) \emph{without} affecting the others, improving stability and predictability. However,, a systematic study of agent-specific post-training is orthogonal to our core contributions and is left to future work.

\subsection{Strategic Search with Tree Memory}\label{sec:search}
Prior LLM-driven kernel optimizers typically use either \emph{best-of-$K$} sampling that generates multiple candidates independently and select the fastest correct one or \emph{iterative refinement} which repeatedly edits the latest kernel~\citep{ouyang2025kernelbench}. However, best-of-$K$ is unguided and wasteful: all the new attempts ignore feedback from earlier attempts and repeatedly probe redundant regions of the design space. On the other hand, iterative refinement is feedback-aware but \emph{myopic}: by building only on the most recent candidate, it is prone to getting trapped in narrow, suboptimal basins.

To address these limitations, \stark reframes kernel optimization as \textbf{strategic search} over a persistent \textbf{tree memory}. We maintain a search tree $T$ whose nodes store candidates and their observations (runtime, correctness, and compiler diagnostics). The root represents the source architecture; each edge corresponds to applying a grounded instruction from the plan agent and realizing it via the code agent (or repairing via the debug agent). Each node $n$ is assigned a score $s(n)$ reflecting competitiveness; in our implementation we use the straightforward kernel runtime as $s(n)$ and treat \emph{lower is better}. For kernels that are incorrect or failing to compile, we give them scores of $+\infty$.
At each step, we (1) \emph{select} a node to expand using a strategic policy, (2) \emph{expand} by invoking the plan/code (or debug) agents to produce a child candidate, (3) \emph{evaluate} for correctness and runtime, and (4) \emph{record} results in $T$ to inform subsequent selections. This converts ad-hoc trial-and-error into a directed, feedback-driven process.

\textbf{Policy choice and an adapted $\epsilon$-greedy rule.}
We compared representative search policies including Monte-Carlo Tree-Search (MCTS), evolutionary, greedy, and $\epsilon$-greedy policies and found that $\epsilon$-greedy consistently performs best under the same budget constraint.  
Importantly, we observe that kernel optimization poses domain-specific challenges that are root dominance (it is very challenging to even outperform the source architecture in the root node) and frequent compilation/runtime failures. 
To address these challenges, we adapt the canonical rule as follows: (1) \textbf{Root throttling:} cap the number of direct children of the root at $n_{\text{root}}$ to avoid redundant first-hop edits; once the cap is reached, the root is ineligible for selection; (2) \textbf{Dead-branch pruning:} if a node has more than $n_{\text{child}}$ children and all current children fail, mark the node ineligible to prevent wasting trials; (3) \textbf{High exploration rate:} use a relatively large $\epsilon$ (empirically $0.3$--$0.4$) to counteract local traps; (4) \textbf{Leaf-biased exploration:} with probability $\epsilon$, sample uniformly from expandable leaves (not only failing nodes), encouraging discovery beyond the immediate failure set.

\subsection{Grounded Instruction}\label{sec:grounded_instruction}
We introduce grounded instruction for kernel enhancement. The plan agent must not only propose an optimization, but also insert \textbf{explicit span anchors} in the kernel source that mark exactly where the change should occur. Each anchor is a short, machine-checkable tag (i.e, \verb|<<<IMPROVE BEGINS>>>| \dots\ \verb|<<<IMPROVE ENDS>>>|) wrapped around the target site, such as a load/store, loop body, or the launch configuration. The code agent consumes this annotated scaffold and resolves each anchor by emitting concrete CUDA that realizes the instruction. Grounded instruction tightens plan--code alignment, curbs hallucinated guidance, and narrows the coder's search space. It also improves traceability: every proposal leaves a visible, verifiable footprint in the final code.
In practice, we observe fewer misinterpretations and markedly fewer faulty kernels. Despite its simplicity, the mechanism is especially effective on Level~3 KernelBench tasks with deeper architectures (e.g., VGG).

\subsection{Dynamic Context Window}
\label{sec:context_window}

\begin{figure}[h]
    \centering
    \includegraphics[width=0.9\linewidth]{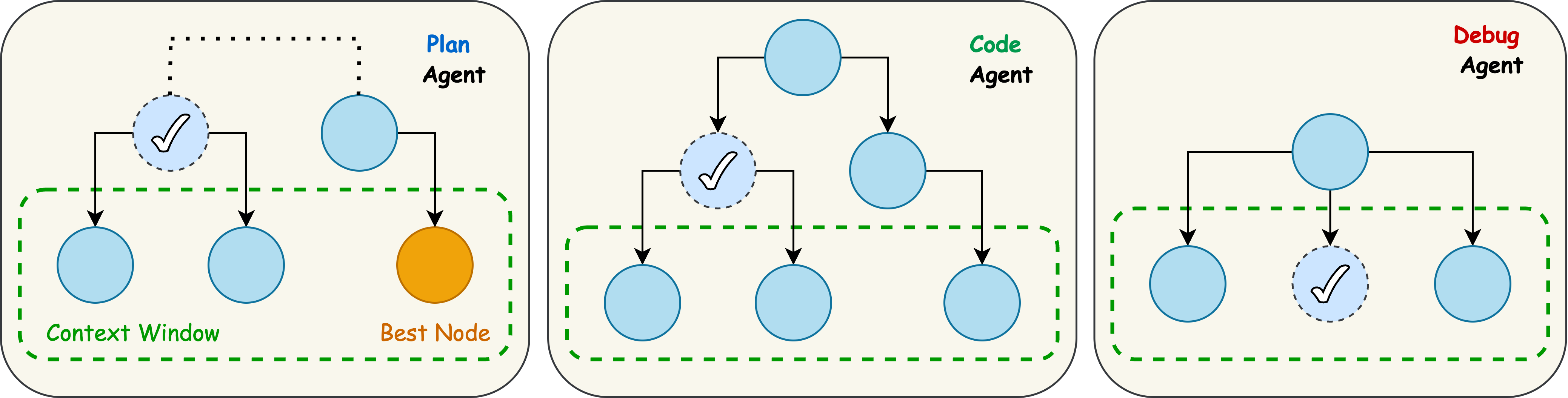}
    \caption{Dynamic Context Window. Nodes with $\checkmark$'s represent selected nodes.}
    \label{fig:dynamic_context_window}
\end{figure}

Past attempts provide rich, actionable signals for subsequent decisions, but different agents benefit from different \emph{views} of this history. We therefore maintain a \emph{dynamic, agent-specific context window} that is rebuilt at each selection step for different agents. 
See Figure~\ref{fig:dynamic_context_window} for a visual demonstration.
Throughout this section, let node $i$ be the node selected by the search policy defined in Section~\ref{sec:search}. We use $\mathcal{W}(i)$ to denote the context window containing a subset of historical attempts and their evaluation outcomes (e.g., compiler information and runtime). As we always include the source architecture as part of the prompt for agents,  $\mathcal{W}(i)$ always includes the root node $n_{\text{root}}$.   
For a naive search algorithm without dynamic context window, $\mathcal{W}(i)=\{i,n_{\text{root}}\}$ only includes node $i$ in addition to the root.
% \stark constructs different sets of historical attempts (nodes) for different agents.

\textbf{Tree relations.} We use tree relations to build agent-specific context windows.
Let $p(i)$ be the parent of node $i$. Define the \emph{siblings} of $i$ as
$\mathcal{S}(i)=\{\,j:\,p(j)=p(i)\,\}$. Moreover, define the set of child nodes of a node $i$ as $\mathcal{D}(i)$. 
We also maintain a small global leaderboard $\mathcal{C}$ of top-performing nodes. 

% \textbf{Plan agent.}
% To facilitate planning, we include all of node $i$'s children into the context window. This brings several benefits. First of all, the plan agent can reflect on its historical instructions to improve the same implementation and find a better optimization strategy.
% Meanwhile, we observe that a common failure mode of LLMs optimizing GPU kernels is that they fail to implement the planned optimization strategy.
% To this end, the plan agent can understand the capability of the code agent by checking how its historical instructions were implemented and the successful and failed implementations from the code agent.
% Importantly, we enforce the plan agent to adjust its instruction based on the code agent's historical implementations. This can significantly enhance the success rate of optimization, preventing the code agent from stucking into implementing instructions beyond its ability.

% In addition, we add include into $\mathcal{W}(i)$ a small set of global competitors from the leaderboard $\mathcal{C}$. This calibrates ambition and prevents redundant exploration: the planner can borrow motifs from winning kernels (e.g., warp-shuffle reductions, vectorized LD/ST, shared-memory tiling) and avoid paths already dominated elsewhere. 

% Thus, we have 
% \[
% \mathcal{W}_{\text{plan}}(i) \;=\; \mathcal{D}(i)\;\cup\;\text{Top-}r(\mathcal{C}).
% \]

\textbf{Plan agent (local \& contrastive global context).}
For a selected node $i$, the plan agent conditions on a context window $\mathcal{W}_{\text{plan}}(i)$ that aggregates node $i$'s children and  a small set of global leaders from the leaderboard $\mathcal{C}$. Formally,
\[
\mathcal{W}_{\text{plan}}(i)\;=\;\{i,n_{\text{root}}\}\;\cup\;\mathcal{D}(i)\;\cup\;\mathrm{Top}\!-\!r(\mathcal{C}),
\]
where $\mathcal{D}(i)$ contains all evaluated children of $i$ with their observations, and $\mathrm{Top}\!-\!r(\mathcal{C})$ returns the $r$ highest-scoring distinct kernels from the global leaderboard (excluding $i$'s subtree) to discourage duplication.

This design serves three purposes. (i) \textbf{reflection}: the plan agent can revise or stack its prior instructions rather than rediscovering them; (ii) \textbf{ambition calibration}: top competitors prevent redundant exploration and provide transferable motifs such as warp-shuffle reductions, vectorized LD/ST, and shared-memory tiling; (ii) \textbf{capability estimation}: by inspecting how past instructions were realized or failed by the code agent, the next instruction is adapted to what the code agent can reliably execute, \emph{improving first-pass success and avoiding instructions beyond current ability}. To achieve this, we explicitly require the plan agent to adapt its instruction to the code agent's demonstrated capabilities observed in $\mathcal{D}(i)$.

\textbf{Code agent (extended context).}
For kernel code emission at node $i$, the code agent conditions on
\[
\mathcal{W}_{\text{code}}(i) \;=\; \{i,n_{\text{root}}\}\;\cup\;\mathcal{D}(i)\;\cup\;\{j:p(j)\in S(i)\}.
\]
The nodes in $\{j:p(j)\in S(i)\}$ are essentially the children of node $i$'s siblings.
Our insight is that these nodes typically share near-identical scaffolds with node $i$ from a common planning lineage, so successful patches and micro-optimizations transfer with high probability; conversely, seeing failures in closely related contexts helps the coder avoid repeating the same mistakes.
Hence, this extended window serves two aims: (i) \textbf{reduce implementation errors} by letting the coder imitate successful patches from closely related scaffolds and avoid previously observed failure modes; (ii) \textbf{surface stronger implementations} by transferring micro-optimizations (e.g., warp-shuffle motifs, vectorized LD/ST, shared-memory tiling) that have already worked on cousin nodes. 

\textbf{Debug agent (local context).} For fault repair, we construct the context window for the debug code as 
\[
\mathcal{W}_{\text{debug}}(i) \;=\; \{i,n_{\text{root}}\}\;\cup\;\mathcal{S}(i),
\]
We choose this design mainly for two reasons. Most fixes are structural and local, e.g., off-by-one guards, stride/indexing alignment, launch-parameter tweaks, or shared-memory sizing often transfer directly among siblings that share the same scaffold. Moreover, restricting the window to $\mathcal{S}$ avoids distracting the debug agent with globally unrelated kernels, improving precision and reducing hallucinated edits.

\subsection{Framework Overview}\label{sec:overview}

Here we provide an overview of our framework \stark and describe its execution process. Algorithm~\ref{alg:stark} presents its pseudocode.

At a high level, \stark repeatedly (i) selects a promising node (a prior attempt) from a search tree, (ii) builds agent-specific context windows from local history and global leaders, (iii) asks the \emph{planning agent} to propose a concrete optimization along with \emph{grounded instruction} anchors inserted into a scaffold, (iv) asks the \emph{code agent} to realize those anchors into an executable kernel, (v) if the selected node has a problematic kernel, we build the debugger’s dynamic context window and request a minimal fix. The new attempt is evaluated, appended as a child node, and the leaderboard $\mathcal{C}$ is updated. We repeat this process until we reach a pre-specified max attempts $B$.

\begin{algorithm}[h]
\caption{STARK: Strategic Team of Agents for Refining Kernels}
\label{alg:stark}
\begin{algorithmic}[1]
\Require Budget $B$ (max attempts), selection policy $\pi_{\text{select}}$ (adapted $\varepsilon$-greedy), leaderboard size $r$
\State Initialize search tree $T$ with root $n_{\text{root}}$ (PyTorch reference)
\State Initialize leaderboard $\mathcal{C} \leftarrow \{n_{\text{root}}\}$
\For{$t = 1,2,\dots,B$}
  \State $i \leftarrow \pi_{\text{select}}(T,\mathcal{C})$ \Comment{pick a node to refine}
  \If{$\textsc{HasBug}(i)$} \Comment{compile fail or unit-test fail recorded at $i$}
    \State $\mathcal{W}_{\text{debug}}(i) \leftarrow \textsc{BuildContextDebug}(i, T)$
    \State $\textsf{kernel}' \leftarrow \textsc{DebugAgent}(\mathcal{W}_{\text{dbg}},\, i.\textsf{kernel},\, i.\textsf{logs})$
    \State $(\textsf{ok},\,\textsf{correct},\,\textsf{runtime},\,\textsf{logs})
           \leftarrow \textsc{Evaluate}(\textsf{kernel}')$ \Comment{compile, correctness check, timing}
    \State $(\textsf{plan},\textsf{anchors}) \leftarrow (i.\textsf{plans},i.\textsf{anchors})$
  \Else
    \State $\mathcal{W}_{\text{plan}}(i) \leftarrow \textsc{BuildContextPlan}(i, T, \mathcal{C})$
    \State $(\textsf{plan},\,\textsf{anchors}) \leftarrow \textsc{PlanAgent}(\mathcal{W}_{\text{plan}})$
    \State $\mathcal{W}_{\text{code}}(i) \leftarrow \textsc{BuildContextCode}(i, T)$
    \State $\textsf{kernel}' \leftarrow \textsc{CodeAgent}(\mathcal{W}_{\text{code}},\, \textsf{plan},\, \textsf{anchors})$
    \State $(\textsf{ok},\,\textsf{correct},\,\textsf{runtime},\,\textsf{logs})
           \leftarrow \textsc{Evaluate}(\textsf{kernel}')$
  \EndIf
  \State $j \leftarrow \textsc{AddChild}(T,\, i,\, \textsf{kernel}',\, \textsf{plan},\, \textsf{anchors},\, \textsf{ok},\, \textsf{correct},\, \textsf{runtime},\, \textsf{logs})$
  \State $\mathcal{C} \leftarrow \textsc{UpdateLeaders}(\mathcal{C},\, j,\, r)$
\EndFor
\State \Return $\textsc{Best}(\mathcal{C})$ \Comment{fastest correct, grounded kernel}
\end{algorithmic}
\end{algorithm}

\section{Experiments}\label{sec:experiments}
We use KernelBench~\citep{ouyang2025kernelbench}, a recently proposed benchmark consisting of comprehensive and challenging GPU kernel tasks, to validate the effectiveness of our proposed approaches. 

\textbf{Baselines and Metrics.} We compare our framework \stark with the following list of approaches: 
\begin{itemize}
    \item {\bf Torch Eager}: the out-of-box PyTorch modules without any compilation or optimization. 
    \item {\bf Torch Compile }: We use \texttt{torch.compile} to produce optimized versions of the given PyTorch modules.  While \texttt{torch.compile} offers different compilation modes, we compare to two of the most representative and competitive ones -- {\bf default} and {\bf max-autotune}. 
    \item {\bf Sampling Agent}: the single agent framework originally proposed and used by KernelBench to evaluate the difficulty of the tasks in KernelBench and the ability of LLMs to write efficient kernels. This agent repeatedly samples responses when given the source model to optimize and chooses the best generated custom kernel as the solution. 
    \item {\bf Reflexion Agent}: this agent follows the Reflexion paradigm~\citep{shinn2023reflexion}, where at each optimization step, it tries to update its last attempt using its corresponding observations such as the compiler and runtime information. 
    % \item {\bf KernelLLM}: a latest LLM specializing with writing GPU kernels, finetuned with a large amount of paired PyTorch modules and their Triton kernel implementation.
    % \item {\bf \stark}: our framework.
     % \item {\bf \stark w/o search}: our multi-agent framework without advanced components related to search and planning. 
\end{itemize}

We report the following metrics to comprehensively understand the agents' performances: (i) $\mathbf{Fast}_1$ rate is the percentage of the problems for which the agent can generate kernels that are \emph{at least} as fast as the torch baselines; (ii) \textbf{Success} rate represents the percentage of the problems for which the agent can generate compiled and correct kernels; (iii) \textbf{Speed}: To better understand how good the generated kernels are, we also report the average speed across all tasks. 
% \begin{enumerate}
%     % \item \textbf{Pass@K}: For each kernel generation task, we sample $n$ candidates and mark those that are \emph{both} correct and faster than Torch Eager and Torch Compile. The pass@k metric is the probability that at least one of the top-$k$ sampled candidates succeeds, estimated as
%     % \[
%     % \mathrm{pass@}k = 1 - \frac{\binom{n-c}{k}}{\binom{n}{k}},
%     % \]
%     % where $c$ is the number of successful candidates. We report the mean pass@k across all tasks.
%     \item $\mathbf{Fast}_1$ rate: the percentage of the problems for which the agent can generate kernels that are \emph{at least} as fast as the torch baselines.
%     \item \textbf{Success} rate: the percentage of the problems for which the agent can generate compiled and correct kernels.
%     \item \textbf{Speed}: To better understand how good the generated kernels are, we also report the average speed across all tasks. 
%     % \item \textbf{Speedup}: The average speed of \emph{improved} kernels. 
% \end{enumerate}

\paragraph{Comparison with Torch Baselines.} In Table~\ref{tab:kernelbench_perf}, we present the results about success rate, $\mathbf{Fast}_1$ rate and speed over all $3$ levels of KernelBench challenges. For each task, we let all agents to have a maximum of $B=30$ attempts. Due to limited computation resource, we evaluate on the representative subset of KernelBench~\citep{ouyang2025kernelbench}. We use \texttt{Claude Sonnet 4} as the base LLMs for all the LLM-based baselines and our agents. Due to space constraint, we defer implementation and evaluation details to Appendix~\ref{app:implementation-details}.

The results in Table~\ref{tab:kernelbench_perf} demonstrate that our proposed framework, \stark, consistently outperforms both the Sampling and Reflexion baselines across all KernelBench difficulty levels. At Level~1, \stark not only achieves a perfect $100\%$ success rate but also delivers up to a $3.0\times$ speedup over Torch Eager baselines, while Sampling and Reflexion agents frequently generate kernels that are slower than the baselines. This advantage becomes even more pronounced at Level~2, where the complexity of the kernels increases. Here, \stark maintains a perfect success rate and achieves speedups of $2.7\times$, whereas the Reflexion agent, despite attaining $100\%$ correctness, produces kernels that run slower than the baseline. At Level~3, which involves the most irregular and challenging tasks, both Sampling and Reflexion degrade significantly, with success rates falling and runtimes dropping below baseline. In contrast, \stark continues to maintain full success while producing kernels that outperform the Torch implementations by up to $1.6\times$. These results highlight that \stark not only generates correct kernels but also delivers substantial performance improvements, even as task difficulty increases.

\begin{table}[h]
\centering
\begin{tabular}{l| c | c c | c c | c c|}
\toprule
 \multicolumn{2}{c}{} & \multicolumn{2}{|c|}{\textbf{Torch Eager }} & \multicolumn{2}{c|}{\textbf{Default }} & \multicolumn{2}{c|}{\textbf{Max-autotune}}\\
% \cmidrule(lr){2-2} \cmidrule(lr){3-4} \cmidrule(lr){5-6} \cmidrule(lr){7-8}
\midrule
\textbf{Level $1$} & \textbf{Success $\uparrow$} & $\mathbf{Fast}_1 \uparrow$ & \textbf{Speed} $\uparrow$ & $\mathbf{Fast}_1\uparrow$ & \textbf{Speed}$\uparrow$ & $\mathbf{Fast}_1\uparrow$ & \textbf{Speed}$\uparrow$ \\
\midrule
Sampling Agent & $57.1\%$ & $14.3\%$ & $0.81\times$  & $7.1\%$  & $0.46\times$ & $7.1\%$ & $0.81\times$ \\
Reflexion Agent & $92.6\%$ & $28.6\%$ & $1.24\times$  & $14.3\%$ & $0.57\times$ & $35.7\%$ & $0.92\times$  \\
% KernelLLM &- & - & - & - & - & - & - \\
\stark & $100\%$ & $71.4\%$ & $3.03\times$ & $78.6\%$ & $2.37\times$ & $78.6\%$ & $2.76\times$ \\
\midrule
\textbf{Level $2$} & \textbf{Success} & $\mathbf{Fast}_1$ & \textbf{Speed} & $\mathbf{Fast}_1$ & \textbf{Speed} & $\mathbf{Fast}_1$ & \textbf{Speed} \\
\midrule
Sampling Agent &$87.5\%$ & $50\%$ & $1.06\times$ & $37.5\%$ & $0.91\times$ & $37.5\%$ & $0.91\times$ \\
Reflexion Agent &$100\%$ & $75\%$ & $0.88\times$  & $62.5\%$ & $0.78\times$ & $62.5\%$ & $0.78\times$ \\
% KernelLLM&- & - & - & - & - & - & - \\
\stark &$100\%$ & $100\%$ & $2.69\times$ & $87.5\%$ & $2.51\times$ & $87.5\%$ & $2.52\times$ \\
\midrule 
\textbf{Level $3$} & \textbf{Success} & $\mathbf{Fast}_1$ & \textbf{Speed} & $\mathbf{Fast}_1$ & \textbf{Speed} & $\mathbf{Fast}_1$ & \textbf{Speed} \\
\midrule
Sampling Agent &$100\%$  & $50\%$ & $0.87\times$  & $12.5\%$ & $0.67\times$ & $12.5\%$ & $0.66\times$ \\
Reflexion Agent &$67.5\%$ & $25\%$ & $0.79\times$ & $12.5\%$ & $0.62\times$ &  $12.5\%$ & $0.61\times$ \\
% KernelLLM &- & - & - & - & - & - & - \\
\stark &$100\%$ & $87.5\%$ & $1.58\times$ & $87.5\%$ & $1.27\times$ & $87.5\%$ & $1.26\times$ \\
\bottomrule
\end{tabular}
\caption{Performance of LLM Agents on the KernelBench Tasks. $\mathbf{Fast}_1$ represents the percentage of problems for which the agent can generate custom kernels that are correct and as fast as the Torch baselines (higher is better). 
Speed is computed as the ratio of the kernel runtime of the baseline to that of the generated kernel.}
\label{tab:kernelbench_perf}
\end{table}

\paragraph{Comparison between Agents.} We investigate deeper into the behavior of our agent \stark with the two baseline agents to better understand their optimization behaviors. A deeper analysis of compile and correctness rates, shown in Table~\ref{tab:kernelbench_compile_correct}, provides further insight into why \stark succeeds where baselines struggle. While all agents achieve relatively high compile rates (mostly above $80\%$), the fraction of kernels that are both compilable and correct varies widely. The Sampling agent, for example, compiles over $90\%$ of its outputs on Level~1 but only $43\%$ of these are functionally correct. Reflexion improves correctness slightly through iterative refinement, but its correctness rate remains below $55\%$ at all levels. In contrast, \stark achieves the highest correctness rates across the board, reaching $61.2\%$ on Level~2 tasks. This suggests that \stark’s structured planning and feedback-driven refinement not only increase the chance of generating efficient kernels but also reduce wasted attempts on invalid or incorrect code.
Finally, Figure~\ref{fig:agent_speedups} highlights the dramatic runtime improvements of \stark relative to baseline agents. On Level~1 tasks, \stark achieves over a $10\times$ speedup compared to Sampling and a $13.7\times$ speedup over Reflexion. On Level~2, these gains rise as high as $16\times$, and even at the most challenging Level~3 tasks \stark maintains $5$--$6\times$ improvements. These relative gains indicate that while baselines occasionally achieve correctness, they rarely deliver true runtime efficiency. By contrast, \stark’s ability to jointly optimize for correctness and speed allows it to close both gaps simultaneously. Taken together, these findings confirm that multi-agent collaboration and strategic search are key enablers for scaling LLMs to the demands of GPU kernel optimization.

\begin{table}[h]
\centering
\begin{tabular}{l c c c | c c c}
\toprule
& \multicolumn{3}{c}{\textbf{Compile Rate}$\uparrow$} & \multicolumn{3}{c}{\textbf{Correct Rate}$\uparrow$} \\
\cmidrule(lr){2-4} \cmidrule(lr){5-7}
\textbf{KernelBench Level} & 1 & 2 & 3 & 1 & 2 & 3 \\
\midrule
Sampling Agent & $90.8\%$ & $97.0\%$ & $84.9\%$  & $43\%$ & $44.0\%$ & $15.1\%$\\
Reflexion Agent & $86.0\%$ & $86.2\%$ & $78.9\%$ & $48.3\%$ & $53.4\%$ & $28.4\%$ \\
\stark & $84.5\%$ & $90.7\%$ & $83.4\%$ & $50.6\%$ & $61.2\%$ & $35.5\%$ \\
\bottomrule
\end{tabular}
\caption{Percentages of Successfully Compiled and Correct Kernels.}
\label{tab:kernelbench_compile_correct}
\end{table}

\textbf{Ablations.} We ablate the agentic components of our system. We compare (i) \textbf{Search Agent}, which is a single-agent model equipped with our strategic search, and (ii) \textbf{MA-only}, which employs the multi-agent workflow (plan/code/debug with grounded instruction and dynamic context) using best-of-$K$ sampling instead of search. As shown in Table~\ref{tab:ablation}, both variants outperform the \emph{Sampling} baseline, confirming that each component helps. When combined in \stark, the effects compound: strategic search exploits the structured proposals produced by the multi-agent workflow, yielding the largest gains.

\begin{table}[H]
\centering
\begin{tabular}{l| c c | c c | c c|}
\toprule
 & \multicolumn{2}{|c|}{\textbf{Torch Eager }} & \multicolumn{2}{c|}{\textbf{Default }} & \multicolumn{2}{c|}{\textbf{Max-autotune}}\\
% \cmidrule(lr){2-2} \cmidrule(lr){3-4} \cmidrule(lr){5-6} \cmidrule(lr){7-8}
\midrule
  & $\mathbf{Fast}_1\uparrow$ & \textbf{Speed}$\uparrow$ & $\mathbf{Fast}_1\uparrow$ & \textbf{Speed}$\uparrow$ & $\mathbf{Fast}_1\uparrow$ & \textbf{Speed}$\uparrow$ \\
\midrule
Sampling Agent   & $50\%$ & $0.87\times$  & $12.5\%$ & $0.67\times$ & $12.5\%$ & $0.66\times$ \\
Search Agent  & $67.5\%$ & $0.89\times$ & $25\%$ & $0.71\times$ & $25\%$ & $0.70\times$ \\
MA-Only  & $67.5\%$ & $1.11\times$ & $25\%$ & $0.92\times$ & $25\%$ & $0.91\times$ \\
\stark  & {$87.5\%$} & $1.58\times$ & $87.5\%$ & $1.27\times$ & $87.5\%$ & $1.26\times$ \\
\bottomrule
\end{tabular}
\caption{Ablation on the Proposed Agentic Features.}
\label{tab:ablation}
\end{table}
%  Aio + bok
% Aio + search (last attempt)
% Grounded + dynamic window + bok
% Grounded + dynamic window + search

\section{Conclusion}

In this work, we introduced an agentic framework for GPU kernel optimization that combines multi-agent role play, dynamic context management, and strategic search. 
% By decomposing the kernel optimization process into specialized roles, our approach mirrors the workflow of human engineers while scaling it through the generative and reasoning capabilities of LLMs. The integration of strategic search and context control further enables efficient exploration of the vast kernel design space, balancing creativity with precision.
Our evaluation on KernelBench demonstrated that the proposed framework consistently outperforms baseline methods in both success rate and runtime efficiency, across tasks of varying complexity. These results highlight the value of moving beyond single-agent or unguided sampling approaches, and point to the promise of collaborative, feedback-driven optimization. 
Looking forward, we envision that agentic LLM frameworks will play an increasingly important role in automated system optimization. Extending our approach to broader classes of operators, diverse hardware architectures, and cross-kernel scheduling decisions are natural directions for future research. More broadly, our work suggests that multi-agent LLMs can meaningfully accelerate the co-design of AI algorithms and infrastructure, pushing the boundaries of what is possible in efficient large-scale computation.

% \clearpage
% \newpage
\bibliographystyle{assets/plainnat}
\bibliography{paper}

\clearpage
\newpage
\beginappendix

\section{Implementation and Evaluation}\label{app:implementation-details}

\textbf{Agent Implementation.} We use \texttt{Claude~Sonnet~4} as the base LLMs for all agent baselines and agents of \stark. For both sampling and reflexion agents, we follow KernelBench to set temperature $\tau=0.7$ during generating, with other generation parameters such as top-p set to the default value. For \stark, we use \texttt{Claude Sonnet 4} with temperature $\tau=0.8$ for the plan agent, and $\tau=0.1$ for the code and debug agents. For all tasks, all agent baselines and \stark have a maximum of $B=30$ attempts to optimize each task. Regarding the hyperparameters of \stark, we choose the root throtting number to be $5$, dead-branch pruning number to be $3$, $\epsilon=0.3$ for the search, $r=2$ for the leaderboard $\mathcal{C}$. To prevent exploding context to the LLMs, we set an upper limit for the number of nodes in the dynamic context window: whenever the dynamic context window has more than $5$ nodes, we randomly sample $5$ from all the nodes in the window. We implement \stark with LangGraph~\citep{langgraph}. 

\textbf{Runtime Evaluation.} We evaluate all the Pytorch baselines and LLM-generated kernels on the same NVIDIA A100 $40$GB GPU. 
We use the source code provided by KernelBench at its official repo\footnote{\url{https://github.com/ScalingIntelligence/KernelBench}} to benchmark the kernels' runtime. In particular, to ensure stable measurement, runtime is measured with CUDA events after warm-up runs using fixed input shapes; we choose a large number of $100$ warm-up runs to ensure accurate measurement.

\section{Prompts}\label{app:prompts}
Our prompts follow the templates of KernelBench~\citep{ouyang2025kernelbench}, which has four components: \emph{system message}, \emph{in-context example}, \emph{architecture source code}, and \emph{instruction}. 

As we have multiple agents in \stark with different tasks, they require different prompts to fulfill their tasks. Specifically, we reuse the system message and in-context example from KernelBench for all agents and include the architecture source code regardless of which node is selected for optimization. To motivate the agents to use the already optimized modules such as cuBLAS, we include an additional instruction in the system message to consider using existing highly optimized kernels. We show the system prompt in Figure~\ref{fig:system_message}, the in-context example in Figures~\ref{fig:in_context_example_1} and~\ref{fig:in_context_example_2}. 
In addition, we include the information within the \emph{dynamic context window} and \emph{role-specific instructions} for different agents. See Figure~\ref{fig:prompt_agent} for the prompt template of \stark. We show the role-specific instruction in Figures~\ref{fig:instruction_plan}, ~\ref{fig:instruction_code}, and~\ref{fig:instruction_debug}. 

\begin{figure}[h]
\centering
\lstset{style=promptstyle}
\begin{minipage}{0.9\linewidth}
\begin{lstlisting}[language={}]
{System Message}

Here's an example to show you the syntax of inline embedding custom CUDA operators in torch: The example given architecture is:
{Example Architecture Source Code}
The example new arch with custom CUDA kernels looks like this:
{Example New Architecture Source Code}

You are given the following architecture:
{Architecture Source Code}

Here is your latest attempt:
{Source Code of the Selected Node}

[Dynamic Context Window] You should use the following observations regarding your historical attempts to provide better implementations: 
- Learn from the failed examples to avoid bugs and write successful kernels. 
- Learn from the successful examples to design improved kernels.

**Kernel Source Code #1**
{Source Code of Historical Attempt}

**Compiler Observation**
{Compiler Log}

**Kernel Execuation Result**
{Runtime or Correctness Error}

**Kernel Source Code #2**
[...skipped]

{Role-specific Instruction}
\end{lstlisting}
\end{minipage}
\caption{Prompt Template for Agents.}
\label{fig:prompt_agent}
\end{figure}

\begin{figure}[h]
\centering
\lstset{style=promptstyle}
\begin{minipage}{0.9\linewidth}
\begin{lstlisting}[language={}]
## System Message 

You are an expert in writing efficient code. 
You write custom CUDA kernels to replace the pytorch operators in the given architecture to get speedups. 

You have complete freedom to choose the set of operators you want to replace. You may make the decision to replace some operators with custom CUDA kernels and leave others unchanged. You may replace multiple operators with custom implementations, consider operator fusion opportunities (combining multiple operators into a single kernel, for example, combining matmul+relu), or algorithmic changes (such as online softmax). You are only limited by your imagination.

You should consider using CUDA's existing highly optimized kernels and operations whenever appropriate. Try building on these optimized blocks and further improve it with your custom kernels. 
\end{lstlisting}
\end{minipage}
\caption{System Message for All Agents.}
\label{fig:system_message}
\end{figure}

\begin{figure}[h]
\centering
\lstset{style=promptstyle}
\begin{minipage}{0.9\linewidth}
\begin{lstlisting}[language={}]
## Instruction

- Optimize the architecture named Model with custom CUDA operators!
- Give explicit and actonable advice to improve the efficiency, in terms of the GPU wall-clock time, of the architecture named Model.

- Give ONE advice of the top priority! Don't over-request.
- Include necessary details such as how to change pointers and indices or how to achieve shared memory tiling so that your advice can be correctly implemented.

- After your advice, modify and return the given source code in the following way:
    - Identify the code block whose efficiency can be improved (that is where your advice should be implemented)
      and mark it with comments '<<<IMPROVE BEGIN>>>' at the beginning and '<<<IMPROVE END>>>' at the end
    - The markers '<<<IMPROVE BEGIN>>>' and '<<<IMPROVE END>>>' should be valid comments for the marked coding language. For example, when marking source code of custom kernels, you need to use comments for the C++ language as '// <<<IMPROVE BEGIN>>>' and '// <<<IMPROVE END>>>'; when marking source code of Python, you should use '## <<<IMPROVE BEGIN>>>' and '## <<<IMPROVE END>>>'
    - Add your advice as comments at the identified code block to help the following agent's implementation
    - There will be another agent focusing on improving the efficiency of the identified code block.
    - Return the complete code block with the identified code block as its subpart.

- You should consider using CUDA's existing highly optimized kernels and operations whenever appropriate. Try building on these optimized blocks and further improve it with your custom kernels.
- When presented with multiple prior attempts, you should consider exploration of more diverse optimization strategies. 
- Pay careful attention to the implementation agent's capability demonstrated from the historical implementations. 
- Adjust your advice accordingly to ensure that it can successfully implement. 
\end{lstlisting}
\end{minipage}
\caption{Instruction for the Plan Agent.}
\label{fig:instruction_plan}
\end{figure}

\begin{figure}[h]
\centering
\lstset{style=promptstyle}
\begin{minipage}{0.9\linewidth}
\begin{lstlisting}[language={}]
## Instruction

Optimize the architecture named Model with custom CUDA operators!

- Think about the given advice from human experts and implement the ones that you believe are correct and you are confident implementing.
- Only focus on the code block marked with '<<<IMPROVE BEGIN>>>' and '<<<IMPROVE END>>>'
- Write custom cuda kernel to replace the pytorch operators within the marked code block to improve its efficiency
- Name your optimized output architecture ModelNew.
- Output the new code in codeblocks.
- Explain your implementation and how you follow the advice.
- Using the given tool to return your final structured answer.

Please generate real code, NOT pseudocode, make sure the code compiles and is fully functional.

NO testing code!

\end{lstlisting}
\end{minipage}
\caption{Instruction for the Code Agent.}
\label{fig:instruction_code}
\end{figure}

\begin{figure}[h]
\centering
\lstset{style=promptstyle}
\begin{minipage}{0.9\linewidth}
\begin{lstlisting}[language={}]
## Instruction

Fix the issues of your implementation named ModelNew, which should improve efficiency of the source model named Model.

- ModelNew and Model should have the same functionality, that is, the same input-output mapping.
- The given architecture ModelNew ieither does not compile, or has run-time error, or has different functionality to the source Model.
- Use the given observations to infer bugs and then fix them.
- Explain how the bugs happen and how you fix it.
- Return the fixed bug-free code and name your optimized output architecture ModelNew.

\end{lstlisting}
\end{minipage}
\caption{Instruction for the Debug Agent.}
\label{fig:instruction_debug}
\end{figure}

\begin{figure}[h]
\centering
\lstset{style=promptstyle}
\begin{minipage}{0.9\linewidth}
\begin{lstlisting}[language=Python]
## Example Architecture Source Code 
import torch
import torch.nn as nn
import torch.nn.functional as F


class Model(nn.Module):
    def __init__(self) -> None:
        super().__init__()

    def forward(self, a, b):
        return a + b


def get_inputs():
    # randomly generate input tensors based on the model architecture
    a = torch.randn(1, 128).cuda()
    b = torch.randn(1, 128).cuda()
    return [a, b]


def get_init_inputs():
    # randomly generate tensors required for initialization based on the model architecture
    return []
\end{lstlisting}
\end{minipage}
\caption{In-context Example Architecture for All Agents.}
\label{fig:in_context_example_1}
\end{figure}

\begin{figure}[h]
\centering
\lstset{style=promptstyle}
\begin{minipage}{0.9\linewidth}
\begin{lstlisting}[language=Python]
## Example New Architecture Source Code

import torch
import torch.nn as nn
import torch.nn.functional as F
from torch.utils.cpp_extension import load_inline

# Define the custom CUDA kernel for element-wise addition
elementwise_add_source = """
#include <torch/extension.h>
#include <cuda_runtime.h>

__global__ void elementwise_add_kernel(const float* a, const float* b, float* out, int size) {
    int idx = blockIdx.x * blockDim.x + threadIdx.x;
    if (idx < size) {
        out[idx] = a[idx] + b[idx];
    }
}

torch::Tensor elementwise_add_cuda(torch::Tensor a, torch::Tensor b) {
    auto size = a.numel();
    auto out = torch::zeros_like(a);

    const int block_size = 256;
    const int num_blocks = (size + block_size - 1) / block_size;

    elementwise_add_kernel<<<num_blocks, block_size>>>(a.data_ptr<float>(), b.data_ptr<float>(), out.data_ptr<float>(), size);

    return out;
}
"""

elementwise_add_cpp_source = (
    "torch::Tensor elementwise_add_cuda(torch::Tensor a, torch::Tensor b);"
)

# Compile the inline CUDA code for element-wise addition
elementwise_add = load_inline(
    name="elementwise_add",
    cpp_sources=elementwise_add_cpp_source,
    cuda_sources=elementwise_add_source,
    functions=["elementwise_add_cuda"],
    verbose=True,
    extra_cflags=[""],
    extra_ldflags=[""],
)


class ModelNew(nn.Module):
    def __init__(self) -> None:
        super().__init__()
        self.elementwise_add = elementwise_add

    def forward(self, a, b):
        return self.elementwise_add.elementwise_add_cuda(a, b)
\end{lstlisting}
\end{minipage}
\caption{In-context Optimized Example Architecture for All Agents.}
\label{fig:in_context_example_2}
\end{figure}

% \begin{PromptBox}[Kernel engineer prompt]{Kernel Engineer}\label{box:kernel}
% Task: Optimize the CUDA/Triton kernel for throughput.
% - Keep functional parity with reference.
% - Avoid bank conflicts; reduce register pressure.
% Inputs: {GPU=H100, Batch=128, D=4096}
% \end{PromptBox}

% \begin{figure}
% \centering
% \begin{minipage}{0.9\linewidth}
% \begin{UserMsg}
% Translate to French: "Good morning".
% \end{UserMsg}
% \begin{AssistantMsg}
% "Bonjour."
% \end{AssistantMsg}
% \end{minipage}
% \caption{Prompt–response used in the translation task.}
% \label{fig:prompt-translation}
% \end{figure}

\section{Example KernelBench Tasks}\label{app:kb-example}

Here we show three examples of KernelBench tasks, one for each level. See Figures~\ref{fig:kb-example-1},~\ref{fig:kb-example-2}, and~\ref{fig:kb-example-3} for example tasks in Level 1, 2, and 3. We refer interested readers to~\cite{ouyang2025kernelbench} for the complete list.

\begin{figure}[h]
\centering
\begin{minipage}{0.9\linewidth}
\begin{lstlisting}[language=Python]
import torch
import torch.nn as nn

class Model(nn.Module):
    """
    Simple model that performs a LogSoftmax activation.
    """
    def __init__(self, dim: int = 1):
        super(Model, self).__init__()
        self.dim = dim
    
    def forward(self, x: torch.Tensor) -> torch.Tensor:
        """
        Applies LogSoftmax activation to the input tensor.

        Args:
            x (torch.Tensor): Input tensor of shape (batch_size, dim).

        Returns:
            torch.Tensor: Output tensor with LogSoftmax applied, same shape as input.
        """
        return torch.log_softmax(x, dim=self.dim)

batch_size = 4096
dim = 393216

def get_inputs():
    x = torch.rand(batch_size, dim)
    return [x]

def get_init_inputs():
    return []  # No special initialization inputs needed
\end{lstlisting}
\end{minipage}
\caption{Example KernelBench Level 1 Task.}
\label{fig:kb-example-1}
\end{figure}

\begin{figure}
\centering
\begin{minipage}{0.9\linewidth}
\begin{lstlisting}[language=Python]
import torch
import torch.nn as nn

class Model(nn.Module):
    """
    Model that performs a matrix multiplication, division, summation, and scaling.
    """
    def __init__(self, input_size, hidden_size, scaling_factor):
        super(Model, self).__init__()
        self.weight = nn.Parameter(torch.randn(hidden_size, input_size))
        self.scaling_factor = scaling_factor

    def forward(self, x):
        """
        Args:
            x (torch.Tensor): Input tensor of shape (batch_size, input_size).
        Returns:
            torch.Tensor: Output tensor of shape (batch_size, hidden_size).
        """
        x = torch.matmul(x, self.weight.T)  # Gemm
        x = x / 2  # Divide
        x = torch.sum(x, dim=1, keepdim=True) # Sum
        x = x * self.scaling_factor  # Scaling
        return x


batch_size   = 1024  
input_size   = 8192  
hidden_size  = 8192 
scaling_factor = 1.5

def get_inputs():
    return [torch.rand(batch_size, input_size)]

def get_init_inputs():
    return [input_size, hidden_size, scaling_factor]
\end{lstlisting}
\end{minipage}
\caption{Example KernelBench Level 2 Task.}
\label{fig:kb-example-2}
\end{figure}

\begin{figure}
\centering
\begin{minipage}{0.9\linewidth}
\begin{lstlisting}[language=Python]
import torch
import torch.nn as nn
import torch.nn.functional as F
import math

class NewGELU(nn.Module):
    """
    Implementation of the GELU activation function currently in Google BERT repo (identical to OpenAI GPT).
    Reference: Gaussian Error Linear Units (GELU) paper: https://arxiv.org/abs/1606.08415
    """
    def __init__(self):
        super(NewGELU, self).__init__()
    
    def forward(self, x):
        return 0.5 * x * (1.0 + torch.tanh(math.sqrt(2.0 / math.pi) * (x + 0.044715 * torch.pow(x, 3.0))))

class CausalSelfAttention(nn.Module):
    """
    A vanilla multi-head masked self-attention layer with a projection at the end.
    It is possible to use torch.nn.MultiheadAttention here but I am including an
    explicit implementation here to show that there is nothing too scary here.
    """

    def __init__(self, n_embd, n_head, attn_pdrop, resid_pdrop, max_seqlen):
        super().__init__()
        assert n_embd % n_head == 0
        # key, query, value projections for all heads, but in a batch
        self.c_attn = nn.Linear(n_embd, 3 * n_embd)
        # output projection
        self.c_proj = nn.Linear(n_embd, n_embd)
        # regularization
        self.attn_dropout = nn.Dropout(attn_pdrop)
        self.resid_dropout = nn.Dropout(resid_pdrop)
        # causal mask to ensure that attention is only applied to the left in the input sequence
        self.register_buffer("bias", torch.tril(torch.ones(max_seqlen, max_seqlen))
                                     .view(1, 1, max_seqlen, max_seqlen))
        self.n_head = n_head
        self.n_embd = n_embd
        [...skipped]
    
class Model(nn.Module):
    """ an unassuming Transformer block """

    [...skipped]

    def forward(self, x):
        x = x + self.attn(self.ln_1(x))
        x = x + self.mlpf(self.ln_2(x))
        return x

[...skipped]
\end{lstlisting}
\end{minipage}
\caption{Example KernelBench Level 3 Task.}
\label{fig:kb-example-3}
\end{figure}

% \section{Post-Training for Coding Experts}

% In our extensive experiments, we observe that a major bottleneck of the existing agent system is code agent's ability in correctly implementing the planning agent's instruction. 

% In fact, one of the most distinctive and contributing features of \stark is its ability to adjust to the coding agent's ability to generate working kernels where baseline agents all fail. Thus, it is an important research question, also with substantive practical value, to ask whether we can improve the coding agent's instruction-following ability in authoring kernels. In particular, we would like to activate the underlying LLMs' ability in writing kernel codes. It would be highly beneficial if we can instantiate the code agent with a moderate sized LLM specially tuned for writing kernels. This can improve both the agent framework's search performance and hardware efficiency. 

% {\bf Post-Training.} We tried a series of moderate-sized LLMs pretrained for code instruct-following (e.g., {\color{blue} Qwen2.5-Coder-7B}). However, we observe that the chance of these models generating correct kernels are too low that post-training through reinforcement learning shows horrible efficiency. 

% {\bf Dataset.} We provide a high-quality dataset, collected from the execution of \stark during our experiments, consisting of the LLMs' correct implementation of the planing agent's instruction. 

\section{Related Work}
\label{app:complete_related_work}

Optimizing GPU kernels to extract maximum performance from underlying hardware is a long-standing and formidable challenge. The vast, non-convex, and hardware-specific search space of possible kernel implementations necessitates sophisticated optimization strategies. The evolution of these strategies can be broadly categorized into three paradigms: empirical auto-tuning, compiler- and model-driven optimization, and most recently, generative approaches using Large Language Models (LLMs). Our work builds upon this trajectory by introducing a fully autonomous agent that manages the entire optimization lifecycle.

\subsection{Empirical and Compiler-Based Optimization}
The foundational approach to GPU performance tuning is empirical auto-tuning, which treats the problem as a black-box search over a set of tunable parameters, such as thread block dimensions, memory tiling factors, and loop unrolling factors~\citep{vanwerkhoven2019kerneltuner}. Traditional methods often rely on an exhaustive or brute-force search, where thousands of potential kernel configurations are generated, compiled, and benchmarked to identify the top performer~\citep{kurzak2012autotuning}.While effective, this process is prohibitively time-consuming; for instance, an exhaustive search for a single GEMM kernel can take over 700 minutes to complete~\citep{nvidia2024nvmatmulheuristics}.

To mitigate this cost, heuristic-driven methods prune the search space. NVIDIA's \texttt{nvMatmulHeuristics}, for example, uses a predictive model to recommend a small subset of high-potential configurations, achieving near-optimal performance in a fraction of the time~\citep{nvidia2024nvmatmulheuristics}. Frameworks like Kernel Tuner~\citep{vanwerkhoven2019kerneltuner}, ATF~\citep{rasch2017atf}, and CLTune~\citep{nugteren2015cltune} provide robust environments for orchestrating these searches and support more advanced strategies like Bayesian Optimization, which builds a probabilistic performance model to guide the search more intelligently~\citep{hellsten2023bacofastportablebayesian, heldens2023kernellauncher}.

Concurrently, compiler-based approaches aim to automate optimization through a series of program transformations applied to an intermediate representation~\citep{yang2010gpgpu}. GPU compilers employ passes for memory coalescing, data prefetching, vectorization, and loop optimizations to adapt naive code to the hardware architecture~\citep{nvidia2008gpuprogram}. While these approaches excel at finding optimal configurations within a predefined search space, they cannot fundamentally alter the kernel's algorithm. Our work introduces an agent that reasons about performance bottlenecks to implement novel, structural code changes, moving beyond simple parameter tuning.

\subsection{Machine Learning for Code Optimization}
Machine learning (ML) has emerged as a powerful tool to transcend the limitations of hand-crafted heuristics. Early work focused on using ML to make better decisions within existing compiler and tuning frameworks. Systems like TVM employ a learned cost model to predict the performance of kernel variants, guiding the search process and avoiding exhaustive empirical evaluation~\citep{chen2018tvm}. More recent efforts have integrated ML directly into production compilers. Google's MLGO framework uses reinforcement learning (RL) to train policies for classic compiler optimizations like function inlining and register allocation, demonstrating significant improvements in code size and performance over decades-old, manually-tuned heuristics in LLVM~\citep{trofin2021mlgo, marinov2024offlineimitationlearningmultiple}. These models can learn from massive code corpora and discover complex feature interactions that are opaque to human experts~\citep{cummins2017endtoend}. 

A more profound application of ML has been in algorithmic discovery. DeepMind's AlphaTensor framed the search for faster matrix multiplication algorithms as a single-player game, using a deep RL agent based on AlphaZero to navigate the enormous search space of tensor decompositions~\citep{fawzi2022alphatensor}. This approach successfully discovered novel, provably correct algorithms that outperform human-derived state-of-the-art methods, including improving upon Strassen's algorithm for $4 \times 4$ matrices for the first time in over 50 years~\citep{fawzi2022alphatensor, deepmind2022alphatensor}. This work marked a critical shift from using ML to \textit{configure} existing optimization strategies to using it to \textit{invent} new ones from first principles. However, AlphaTensor operated in a clean, formal mathematical domain. Translating this power to the messy, syntactic, and hardware-constrained domain of GPU kernel programming presents a distinct challenge. Our work addresses this by employing an agent that operates directly on source code, navigating the complexities of syntax, compilation, and hardware-specific performance characteristics.

\subsection{LLM-Powered Autonomous Agents}
The capabilities of LLMs have given rise to a new paradigm of autonomous agents. An LLM agent uses a core LLM as its "brain" or controller, augmented with capabilities for planning, memory, and tool use to perform complex tasks autonomously~\citep{weng2023llmagents}. The key distinction from simple LLM prompting is the agent's ability to decompose a high-level goal into a sequence of manageable subtasks, execute them iteratively, and use reflection to gauge progress and self-correct. This agentic workflow involves the LLM interacting with an external environment through a set of tools, such as a code interpreter or a web search API, to gather information and perform actions. While this paradigm is powerful for general problem-solving, its application to specialized domains like software engineering requires tailored tools and reasoning processes. Our work specializes this agentic concept for the domain of performance optimization, which presents unique challenges not found in general-purpose agent tasks.

\subsection{LLMs for Code Optimization and Generation}
The advent of powerful LLMs has opened a new frontier in performance engineering. To grant LLMs greater autonomy, the agentic paradigm has been adapted specifically for software engineering. The success of systems like SWE-agent, which autonomously resolves complex bugs in large GitHub repositories, has demonstrated the viability of this approach~\citep{yang2024sweagent}. SWE-agent equips an LLM with a specialized Agent-Computer Interface (ACI) containing tools for file navigation, editing, and test execution, enabling it to perform long-horizon tasks far beyond the scope of simple code generation~\citep{yang2024sweagent, jimenez2024swebench}. While these agents are a significant step towards autonomous software engineering, their focus has primarily been on functional correctness, such as bug fixing. Our work extends this agentic software engineering paradigm to the non-functional, performance-oriented domain of GPU kernel optimization. We introduce an agent that not only interacts with a codebase but also with hardware profiling tools, allowing it to autonomously diagnose performance bottlenecks, form hypotheses, and conduct experiments to iteratively improve kernel efficiency, thus acting as a true autonomous performance engineer.

\end{document}